\documentclass[letterpaper, 10 pt, conference]{ieeeconf}  

\IEEEoverridecommandlockouts                              

\usepackage{booktabs}
\usepackage{multirow}
\usepackage{cite}
\usepackage{todonotes}
\usepackage{comment}
\title{\LARGE \bf
Transferring Multi-Agent Reinforcement Learning Policies for Autonomous Driving using Sim-to-Real 
}

\author{Eduardo Candela*$^{1}$, Leandro Parada*$^{1}$, Luis Marques*$^{1}$, \\
Tiberiu-Andrei Georgescu$^{1}$, Yiannis Demiris$^{2}$, Panagiotis Angeloudis$^{1}$ 
\thanks{*These authors contributed equally.}
\thanks{$^{1}$E. Candela, L. Parada, L. Marques, T. Georgescu and P. Angeloudis are with the Centre for Transport Studies, Department of Civil and Environmental Engineering, Imperial College London, UK
        {\tt\small
        e.candela-garza19@imperial.ac.uk}}%
\thanks{$^{2}$Y. Demiris is with the Personal Robotics Laboratory, Department of Electrical and Electronic Engineering, Imperial College London, UK
}
}

\usepackage{amsmath,amssymb}

\begin{document}

\maketitle
\thispagestyle{empty}
\pagestyle{empty}

\begin{abstract}

Autonomous Driving requires high levels of coordination and collaboration between agents. Achieving effective coordination in multi-agent systems is a difficult task that remains largely unresolved. Multi-Agent Reinforcement Learning has arisen as a powerful method to accomplish this task because it considers the interaction between agents and also allows for decentralized training---which makes it highly scalable. However, transferring policies from simulation to the real world is a big challenge, even for single-agent applications. Multi-agent systems add additional complexities to the Sim-to-Real gap due to agent collaboration and environment synchronization. In this paper, we propose a method to transfer multi-agent autonomous driving policies to the real world. For this, we create a multi-agent environment that imitates the dynamics of the Duckietown multi-robot testbed, and train multi-agent policies using the MAPPO algorithm with different levels of domain randomization. We then transfer the trained policies to the Duckietown testbed and compare the use of the MAPPO algorithm against a traditional rule-based method. We show that the rewards of the transferred policies with MAPPO and domain randomization are, on average, 1.85 times superior to the rule-based method. Moreover, we show that different levels of parameter randomization have a substantial impact on the Sim-to-Real gap.
\end{abstract}

\section{INTRODUCTION}

Coordination of Autonomous Vehicles (AVs) in traffic is a hard problem with a non-trivial optimization objective. On the one hand, rule-based policies can provide acceptable solutions to specific and heavily-simplified situations, but cannot possibly cover all scenarios that can happen in the real world \cite{rear_crashes}, and require manual crafting of the rules. Moreover, rule-based methods cannot achieve effective coordination between agents and cannot adapt to changing environments. On the other hand, multi-agent learning algorithms can achieve effective agent-agent coordination and can potentially explore a great number of different environment configurations. 

Multi-Agent Deep Reinforcement Learning (MARL) is potentially a powerful scalable framework for developing control policies for AVs. MARL uses Deep Neural Networks to represent complex value functions or agents policies that would otherwise require specific hand-crafted rules. However, MARL cannot be trained in live Autonomous Driving systems due to safety concerns \cite{palanisamy2020multi, smarts20} . Furthermore, these algorithms must be trained using thousands or even millions of steps, which cannot be achieved in the real world. Therefore, MARL algorithms must be trained in simulation environments, and then the policies can be transferred to the real system.

   \begin{figure}[t!]
      \centering
      \includegraphics[width=\linewidth]{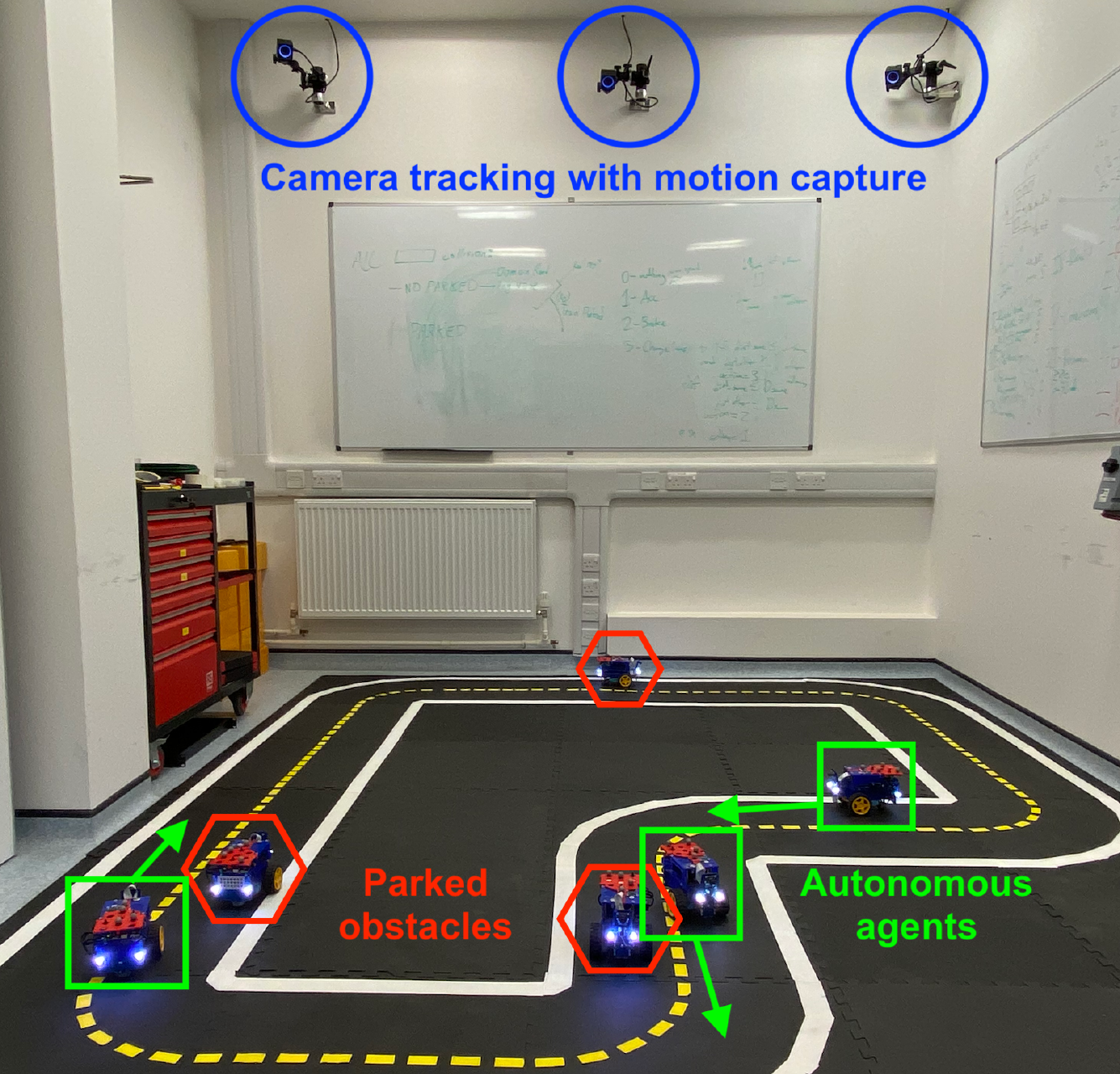}
      \caption{The goal is to train autonomous vehicles to go around a track as fast as possible while avoiding collisions, parked obstacles, and leaving the track. Policies are trained using Multi-Agent Deep Reinforcement Learning in a simulated environment with a kinematic model and different levels of domain randomization that reduce the \textit{sim-to-real gap} by adding noise and uncertainty to the simulations. The trained policies are then transferred to a real fleet of Duckiebot robots, with a motion capture camera system that obtains the vehicles' real pose.}
      \label{fig:lab}
   \end{figure}

Simulations have been an essential tool in the field of reinforcement learning, providing a flexible environment to train and test new algorithmic developments efficiently. However, the performance of policies trained in simulation is not likely to be replicated in the real world. The difference between a simulated environment and its real counterpart is called \textit{reality gap} \cite{DBLP:journals/corr/TobinFRSZA17}, and is usually observed through the overfitting of the learned policy to the unique characteristics of the simulations. 

The \textit{reality gap} can become even larger for multi-agent systems, such as those involving AVs. These systems introduce additional complexities to the Sim-to-Real problem, such as agent collaboration, environment synchronization and limited agent perception. Little work has been done on bridging the gap between simulation and reality in multi-AV systems. Attempts have been made to make MARL more robust to uncertainty \cite{ado19, zhang2020robust}, although these methods have not been tested in the real world. Therefore, in this study, we propose a method to train multi-AV driving policies in simulation and effectively transfer them to reality. For this purpose, we present a multi-agent autonomous driving gym environment that resembles the Duckietown testbed \cite{duckietown17}, which consists of small differential drive robots with two conventional wheels and a passive caster wheel. We then train different multi-agent policies in simulation and transfer them to the Duckietown using Domain Randomization \cite{DBLP:journals/corr/TobinFRSZA17}. We compare the method against a traditional rule-based method on a case study with 3 agents and 3 stationary vehicles which act as obstacles, and show the benefit of training policies with domain randomization when transferring them to reality. 

The rest of this paper is structured as follows. Section \ref{sec:related_work} presents the background and related work. Section \ref{sec:methodology} proposes the methodology including the MARL modeling framework, the Duckietown testbed, the gym environment and the Sim-to-Real transfer mechanism. Section \ref{sec:results} presents the main results and Section \ref{sec:conclusions} concludes with insights to future work.

\section{RELATED WORK} \label{sec:related_work}

\subsection{Multi-Agent Learning for Autonomous Vehicles}

A taxonomy for multi-agent learning in Autonomous Driving was presented in \cite{smarts20}, consisting of five levels, where the first level \textbf{M0} represents rule-based planning with no learning, and the last level \textbf{M5} represents agents with a high degree of forward-planning, working to optimize the \textit{Price of Anarchy} of the overall traffic scenario \cite{poa99,selfish_poa05}. Most multi-agent learning paradigms find it difficult to reach level \textbf{M5}, as most traffic scenarios would be cataloged as massive multi-agent games. Fortunately, a significant amount of algorithms fit either in \textbf{M3}, allowing agents to behave and expect in return partially cooperative behavior \cite{rmaddpg20}, and \textbf{M4}, where a local Nash equilibrium is achieved through the grouping of agents \cite{bi_ac21}.

A few studies have addressed multi-AV problems using MARL. \cite{smarts20} proposed an open-source MARL simulation platform, that includes several traffic scenarios with the possibility of choosing different AV controllers, environment configurations and MARL algorithms. Similarly, \cite{palanisamy2020multi} presented MACAD-gym, an Autonomous Driving multi-agent platform based on the CARLA \cite{carla17} simulator. \cite{chen2021graph} modeled the multi-agent problem as a graph and used Graph Neural Networks in combination with Deep Q Network to control lane-changing decisions in environments with multiple Connected Autonomous Vehicles (CAVs).

\subsection{Sim-to-Real}

To bridge the gap between simulation and reality, known as \textit{reality gap}, various domain-adaptation approaches have been developed \cite{sim2realsurvey20}. To this date, the area with the highest amount of contributions in Sim-to-real is \textit{robotic manipulation}. The methods that have been used include imitation learning, data augmentation and real world reinforcement learning. The latter-most case is the most difficult one to replicate due to lack of resources, safety concerns and difficulty in resetting training runs. There are simple ways of automating the reset in real world robotic manipulation \cite{leavenotrace17}, or to continue training efficiently without resetting \cite{r3l20,resetfree21}. However, in the case of AVs, human interference is needed in order to reset the environment or to prevent catastrophic events. Therefore, most of the techniques used in robotic manipulation are impractical to adapt to environments with AVs.

For AV scenarios, \cite{model21} proposed ModEL, a modular infrastructure which considered perception, planning and control, each being trained using reinforcement learning. They used vision as the agent's main sensorial perception, and the CARLA simulator \cite{carla17} during training for data augmentation and domain generalization, in order to improve overall agent robustness.

Furthermore, in the case of multi-agent systems, there are more reality gaps compared to single-agent settings. Three significant gaps have been reported in \cite{ado19}: \textit{the control architecture} gap, which relates to the tendency of simulators to synchronize the actions of all agents at each time step; \textit{the observation} gap, which relates to the limited perception of agents in scaled-out environments; and \textit{the communication} gap which, similar to the previous one, relates to the highly limited and inconsistent communication which in multi-agent systems. Given traditional methods of software simulation, the aforementioned gaps are non-trivial to overcome, even with a redesign of the simulation itself. The study suggests that more robust modeling of the interaction between agents would be more beneficial. \cite{ado19}  proposed a method called Agent Decentralized Organization (ADO), which encourages agents to share a board of information provided at certain time frames, without the necessity for the information to be complete or up to date.

Although research in Sim-to-Real for AV learning is extensive, the main focus of the literature is generalizability over diverse environments, not different actors. This can make progress in the literature slower, since the hardware is usually significantly different and not open source. Thankfully, there are attempts to standardize the research hardware, with open source sets like Duckietown \cite{duckietown17} and DeepRacer \cite{deepracer19}, which ensure that more prior research becomes easier to reproduce consistently. However, practically no study has focused on transferring multi-agent policies for autonomous driving to the real world.

\section{METHODOLOGY} \label{sec:methodology}

\subsection{Multi-Agent Deep Reinforcement Learning}

The MARL problem can be modeled as a Decentralized Partially Observable MDP (Dec-POMDP), which can be defined by the tuple $ \langle N, S, {A^i}, T, R, \Omega^i, O \rangle  $, where $N$ is a finite set of agents, $S$ is the set of states, $A^i$ is the set of actions for agent $i$, $T:S \times A  \rightarrow S$ is the transition function, $R:S \times A \times S \rightarrow \Re  $ is the reward function, $\Omega^i$ is the set of observations for agent $i$ and $O:S \times A$ is the set of observation probabilities. The goal of Dec-POMPD is to find the joint optimal policy $\pi^*$ that maximizes the expected return.

\subsection{Multi-Agent Proximal Policy Optimization (MAPPO):}

MAPPO \cite{mappo21} is an extension of the Proximal Policy Optimization algorithm to the multi-agent setting. As an on-policy method, it can be less sample efficient than off-policy methods such as MADDPG \cite{lowe2017multi} and QMIX \cite{rashid2018qmix}. Despite this fact, MAPPO has been found to be a strong method for cooperative environments, outperforming state-of-the-art off-policy methods and achieving similar sample efficiency in practice. In addition, knowledge of prior states gets recalled through the usage of recurrent neural networks (RNNs) for the actor and critic networks.

The architecture of MAPPO consists of two separate networks: an actor and a critic network. The actor network uses the observation of each agent, while the critic network observes the whole global state. Therefore, it is directly related to the Centralized Training Decentralized Execution (CTDE) paradigm. The centralized critic can foster cooperation among agents during training, but during execution agents' policies (actors) are implemented in a decentralized way. We assume that all agents share both networks for simplicity, even though MAPPO allows for different networks for each individual agent. 

\subsection{Duckietown testbed}

Duckietown \cite{duckietown17} is a multi-robot testbed that consists of several small autonomous robots (Duckiebots) \cite{db21mdatasheet20} that transit inside a city (Duckietown). Each Duckiebot is a differential drive vehicle equipped with wheel encoders, a monocular camera, an Inertial Measurement Unit (IMU) and a time of flight sensor.

For positioning, we use a NaturalPoint OptiTrack MoCap system, with 8 PrimeX13 cameras tracking at 120 Hz, to obtain the real-time pose of the Duckiebots. A unique asymmetric passive marker configuration is used for each car, and the system was calibrated to be accurate up to 0.5 mm. The 3D pose of each car is broadcasted to the simulator through the NatNet SDK such that position $(x,y)$ and orientation $(\theta)$ are directly obtained after a coordinate transformation, and velocities are calculated from applying finite differences to consecutive measurements.

\subsection{Duckietown Multi-Agent Autonomous Driving Environment (Duckie-MAAD)}

\begin{figure}[t!]
      \centering
      \includegraphics[width=\linewidth]{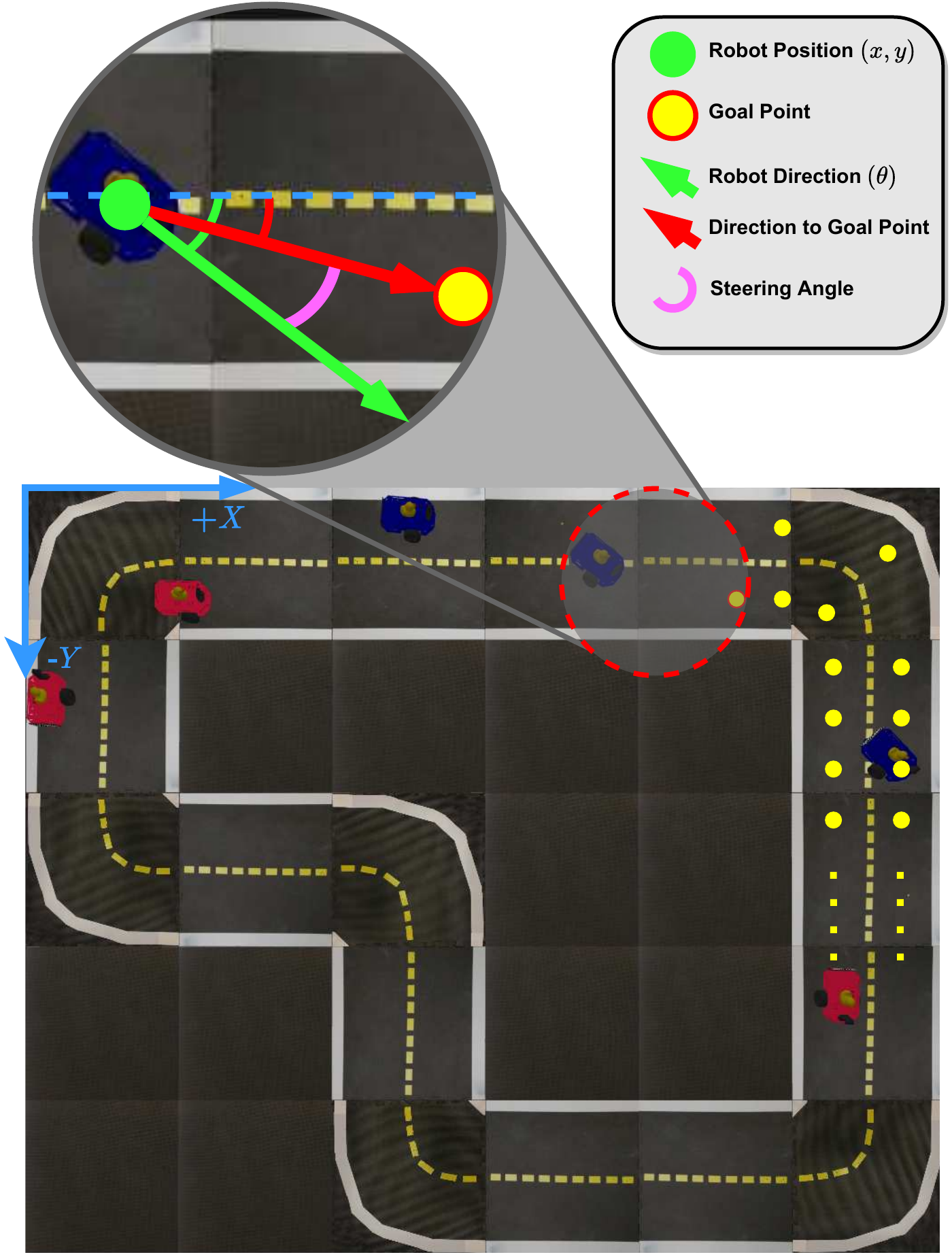}
      \caption{Test track, used in sim and reality, for the Duckie-MAAD gym environment. Yellow circles represent the waypoints along each lane, which extend all the way around the loop. The \textit{goal point} is determined by the Path Following function and the steering angle required to reach it is the difference between the car's direction in world coordinates and the angle between the car's position and the \textit{goal point}.}
      \label{fig:duckie-gym}
  \end{figure}

For the purpose of training and evaluating multi-AV policies, we introduce a new environment called Duckietown Multi-Agent Autonomous Driving Environment (Duckie-MAAD), which is based on the Gym-Duckietown environment \cite{gym_duckietown}. The original Gym-Duckietown is a single-agent lane-following environment that can be used to test different Reinforcement Learning methods. We extend this environment to the multi-agent setting and allow agents to take high-level decisions based on local observations. The agents (Duckiebots) can move around a track, following predefined lane paths and taking decisions to accelerate, brake, or change lanes. In addition, each agent can perceive nearby agents and objects in the track within a given radius. Based on perception, proximity and collision penalties can be assigned to agents, as well as a penalty for leaving the track. Although the agents follow a path predefined by waypoints, they have to learn to stay inside the track, because going too fast will take them off. Figure \ref{fig:duckie-gym} shows a test track with 3 agents (the moving cars) for the Duckie-MAAD environment.

\begin{figure}[t!]
      \centering
      \includegraphics[width=\linewidth]{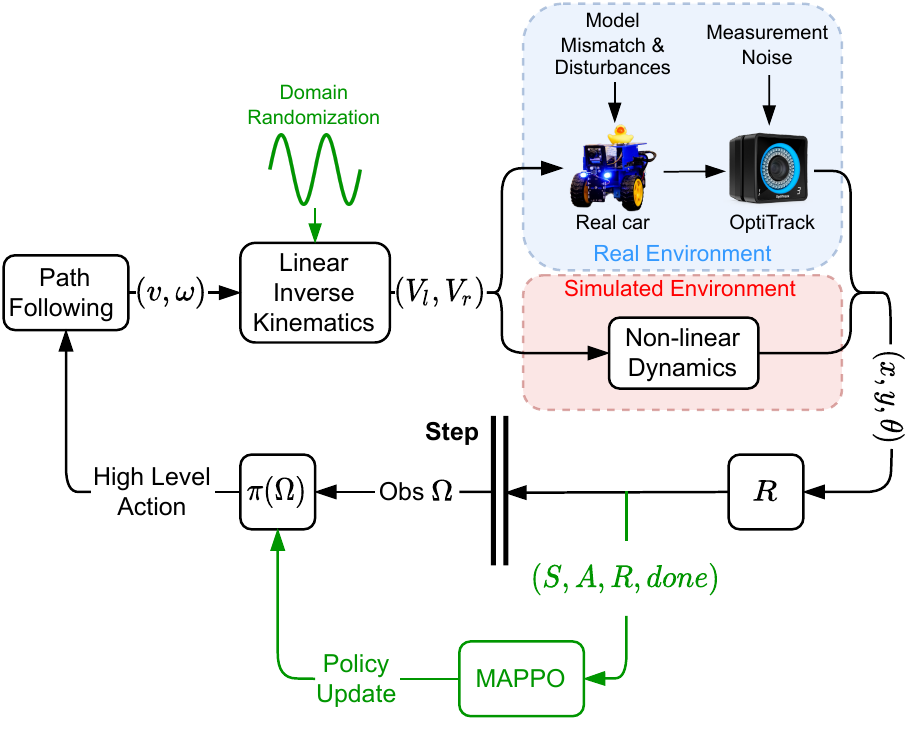}
      \caption{Duckie-MAAD architecture. The step update loop is the following: 1) the agent takes an action following policy $\pi$; 2) using this, the Path Following function selects the next target waypoint and calculates the linear and angular velocities required to reach it; 3) the wheel velocities that lead to such speeds are calculated using (linear) differential drive Inverse Kinematics and sent to the agents (and domain randomization is added if training MARL); 4) the resulting pose for each agent is obtained either from real life or in the simulator, together with a reward R; If training MARL, 5) the policy is updated using MAPPO. The green elements, only when training MARL.}
      \label{fig:architecture-diagram}
   \end{figure}

Upon a new environment step, the MAPPO algorithm takes in the observations of the agents and outputs a high-level action $A$ for each car. Using this as input, a Path Following logic is implemented that determines for each car its next goal waypoint (within a pre-defined list that extends along each lane), and calculates the required linear and angular velocities $(v, \omega)$ to reach it. These are passed to the differential drive Inverse Kinematics models which produces the left and right wheel velocities $(V_l, V_r)$ that lead to this motion. Said velocities are then either fed to the simulated vehicles, where the pose is updated via a Non-linear Dynamics model \cite{ercan2019nonlinear}, or to the real cars, where the pose is then updated via the OptiTrack MoCap system. The new robot poses are then used to calculate the Reward $R$ and the end of of the step is reached. A detailed diagram illustrating this step update in the environment can be found in Figure \ref{fig:architecture-diagram}.


We define the individual elements of the Duckietown Multi-Agent Autonomous Driving Environment as follows:

\begin{itemize}
    \item Agents $N$: Each Duckiebot is treated as an individual agent. All agents are independent, have a local observation of the environment and obtain a unique reward.
    \item Observation $\Omega$: The local observation of each Duckiebot is composed by the steering angle, the distance to the center of the lane, the angle with respect to a tangent to the current lane, local distances to the closest Duckiebots in the same and opposite lane, their respective longitudinal velocities, and a binary variable that shows whether the agent is off the track or not.
    \item Action $A$: A discrete action space with four high-level possible decisions -- accelerate (0.25 $m/s^{2}$); brake (0.25 $m/s^{2}$); change lane; keep previous velocity (no acceleration). In each step agents can accelerate, brake, change lane or do nothing.
    \item Reward function $R$: $v -5c -5t -0.5l$; where $v$ is the agent's measure velocity ($m/s$), and binary variables  $c$ representing a collision, $t$ capturing if the agent is out of the track, and $l$ if the agent is changing lanes.

\end{itemize}

\subsection{Simulation to reality transfer}

There is a significant gap between the simulation environment and reality due to several reasons. First, the agent dynamics within the Duckie-MAAD gym environment are based on a dynamical model, which is based on various assumptions, such as a symmetrical mass distribution of the Duckiebots and that the motors operate in steady-state mode. Second, in the simulator, agent steps are performed sequentially according to the environment's internal clock, meanwhile in reality the agents may take their own step asynchronously. Third, there are certain inaccuracies related to the OptiTrack positioning system. Fourth, although all the Duckiebots share the same architecture, in practice they all behave differently due to small differences in their individual components. All of the above create a significant Sim-to-Real gap that must be addressed in order to successfully implement multi-agent policies trained in simulation.

To tackle this gap, we use domain randomization, which is a technique that consists in training a DRL model over a variety of simulated environments with randomized parameters. The objective is to train a model that can adapt to the real world environment, which is expected to be a sample of the space of environments created through the randomization procedure.

We employ uniform domain randomization of the following parameters within the Duckie-MAAD environment:

\begin{itemize}
    \item Steering factor ($unitless$)
    \item Motor constant ($K$) ($Nm/\sqrt{W}$)
    \item Gain ($unitless$)
    \item Trim ($unitless$)
    \item Steering error ($rad$)
\end{itemize}

\section{RESULTS} \label{sec:results}

To implement and validate the proposed methodology, we made use of the Duckietown Multi-Agent Autonomous Driving Environment for first training and then testing in simulation various policies. Afterward, experiments in real life were conducted following the same goals, rewards and settings. The specific environment utilized consisted of 3 agents and 3 permanently parked cars randomly spawned at the beginning of each episode throughout the test track shown in Figure \ref{fig:duckie-gym}. Agents had the goal to maximize their reward by going around the track as fast as possible while avoiding leaving the bounds of the track and colliding with other cars (either parked or other moving agents). A minimum speed of 0.1 $m/s$ was set for all agents; and a maximum speed of 0.3, 0.4 and 0.5 $m/s$ for each agent, respectively. Both the simulations and real experiments were run at discrete time-steps with a frame rate of 10 Hz. 

We trained three policies using the MAPPO RL method with the following parameters: 

\begin{itemize}
    \item N° episodes for training: $2000$
    \item N° steps per episode: $400$
    \item Learning Rate: $5 \times 10^{-4}$
    \item Critic Learning Rate: $5 \times 10^{-4}$
    \item PPO epochs: $15$
    \item Entropy coefficient: $0.01$
    \item Actor Network: $MLP(64, 64, 4)$
    \item Critic Network: $MLP(64, 64, 1)$
\end{itemize}

\begin{figure}[h]
    \centering
    \includegraphics[width=\linewidth]{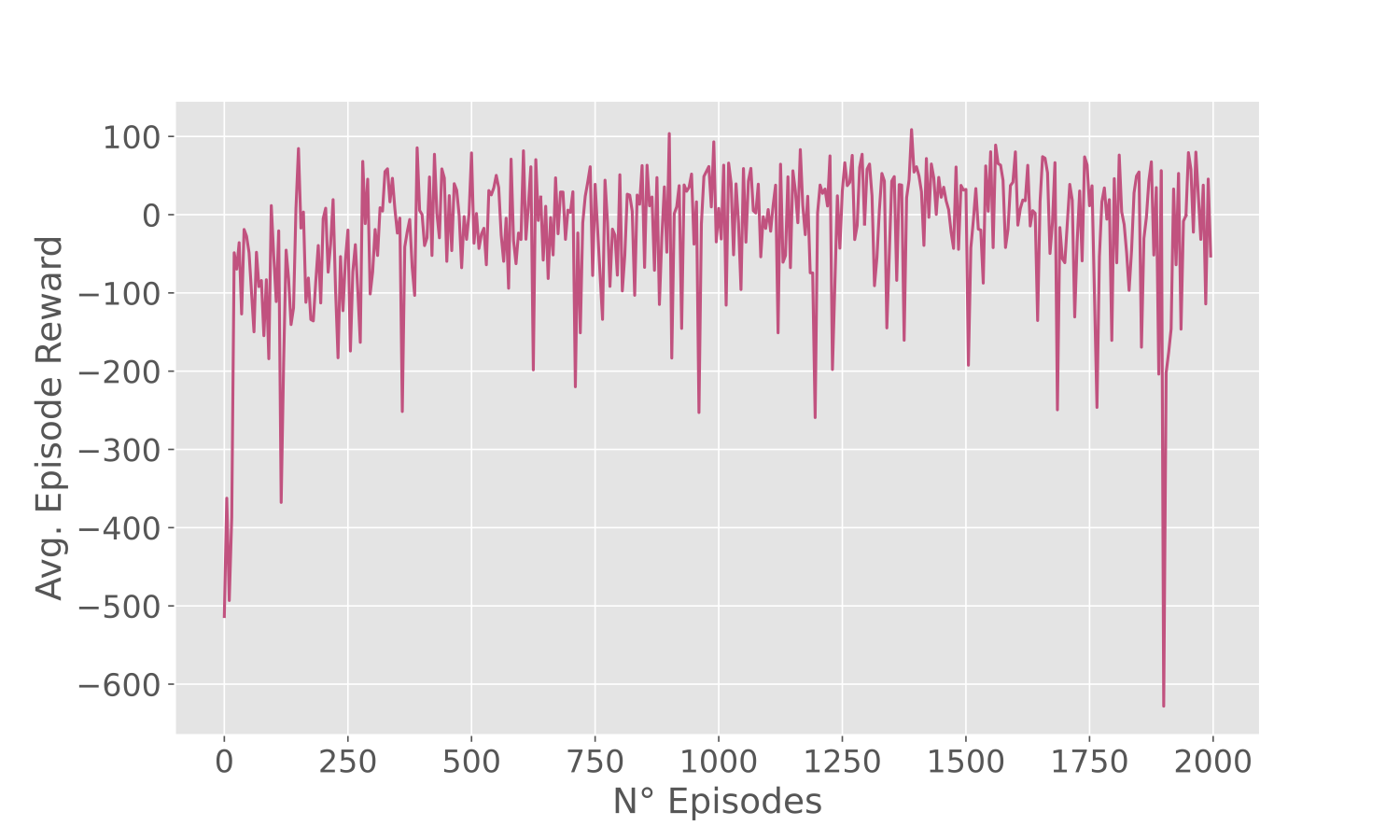}
    \caption{Reward convergence during training of Med D.R. policy.}
    \label{fig:rewards_train}
\end{figure}

Each policy was trained using a different level of domain randomization (none, medium, high) with the distributions shown in Table \ref{table:random_params}. All policies converged after approximately 500 episodes, as can be seen in the reward plot in Figure \ref{fig:rewards_train}. For training, a workstation was used with an AMD Ryzen Threadripper 3990x 64-core processor and a NVIDIA RTX 3090 GPU, which resulted in a training time of approximately 3 hours for each policy.

\begin{table}[h]
\caption{Domain randomization distributions}
\label{table:random_params}
\begin{center}
\begin{tabular}{|l|l|l|l|l|}
\hline
\textbf{Policy} & \textbf{Rule-based} & \textbf{No D.R.} & \textbf{Med D.R.} & \textbf{High D.R.} \\ \hline
Steer factor & 1 & 1 & U(0.8,1.2) & U(0.5,1.5) \\ \hline
$K$            & 27 & 27 & U(22,32) & U(14,40) \\ \hline
Gain         & 1 & 1 & U(0.8,1.2) & U(0.5,1.5) \\ \hline
Trim         & 0 & 0 & U(-0.1,0.1) & U(-0.15,0.15) \\ \hline
Steer error  & 0 & 0 & N(0,0.1) & N(0,0.5) \\ \hline
\end{tabular}
\end{center}
\end{table}

As a baseline, we implemented Gipps' lane changing model \cite{gipps1986model} -- which is a commonly used rule-based algorithm for describing driving behavior -- and set the safe following distances according to the RSS model introduced in \cite{shalev2017formal}.

The four policies were tested first in the simulated environment and then in the Duckietown testbed (real life) to measure and compare their performance and transferability to real life. To achieve statistical significance and reduce the variance of the average of the metrics to be recorded, 30 runs were executed in the simulator and then another 30 in real life for each policy, with all cars being randomly spawned at the beginning of each run. Each run lasted 400 steps, which granted enough time to the agents to go around the track at least once. 

\begin{figure}[h]
    \centering
    \includegraphics[width=\linewidth]{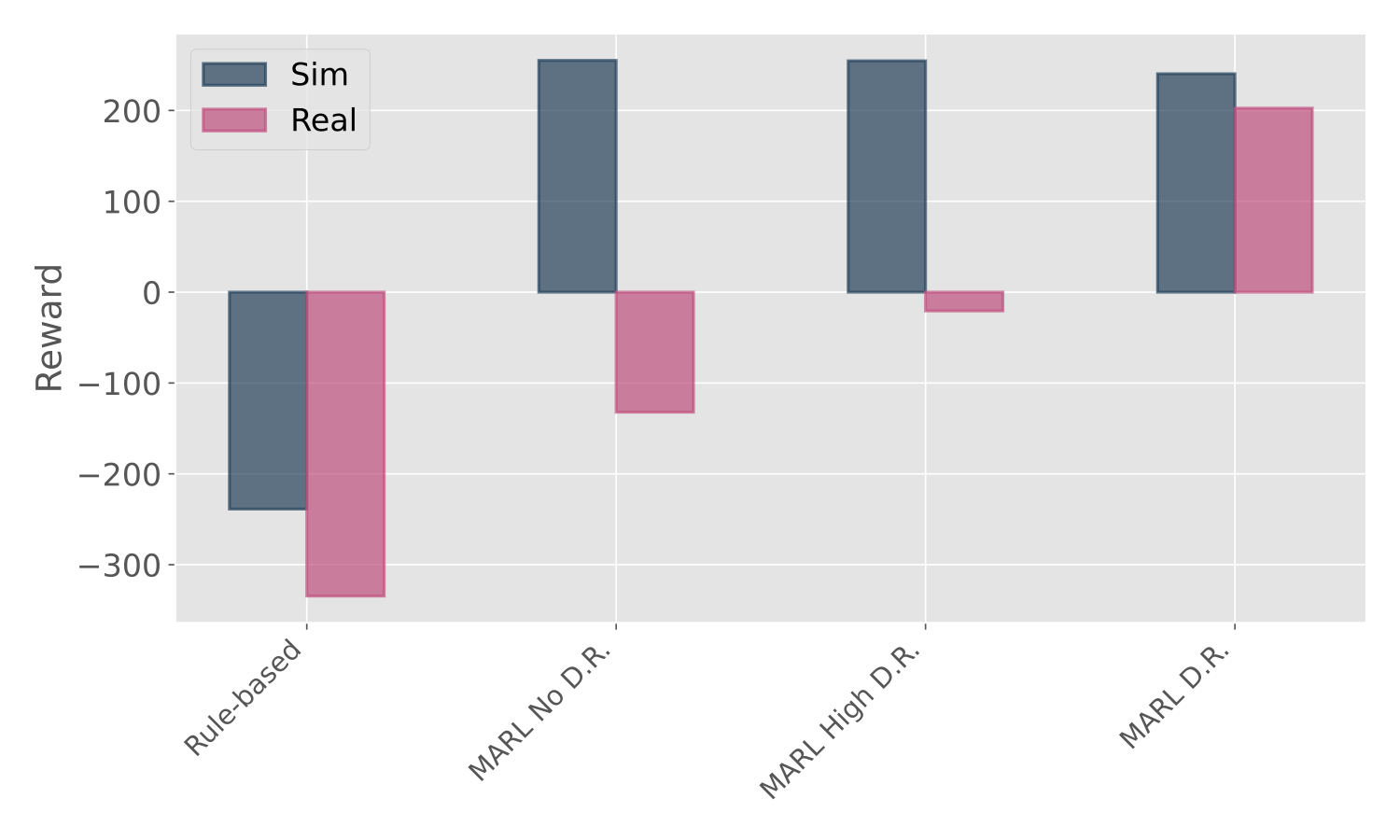}
    \caption{Average rewards for policies in simulation and real life. MARL policies clearly outperform in real life the rule-based baseline, which can be used as a proxy of the reality gap. }
    \label{fig:rewards_comparison}
\end{figure}

The average rewards achieved by each policy in the simulated environment and real life are presented in Figure \ref{fig:rewards_comparison}, where it can be seen that policies on average clearly perform worse in reality than in the simulator. This is due to the previously discussed \textit{reality gap}, which in this case is significant mainly because of the high unreliability of the Duckiebot robots. The rule-based policy performed the worst, as was expected, because despite following safety distances, it cannot adapt to other agents not precisely following their lanes. The policy trained with MARL and no domain randomization achieved the best rewards in simulation, but the worst in real life among the MARL-trained policies. Note that, despite slightly decreasing the simulated performance, the addition of domain randomization improves the rewards in real life. The policy with a medium level of domain randomization performed better than the one with a high level, because the latter learned to be more conservative, as can be appreciated in the following figures. On average, the med D.R. policy provides almost twice as large reward as the rule-based method. The poor performance of the untrained rule-based algorithm can be used as a proxy of the reality gap caused mainly by the unreliability and unpredictability of the real Duckiebots.

\begin{figure}[h]
    \centering
    \includegraphics[width=\linewidth]{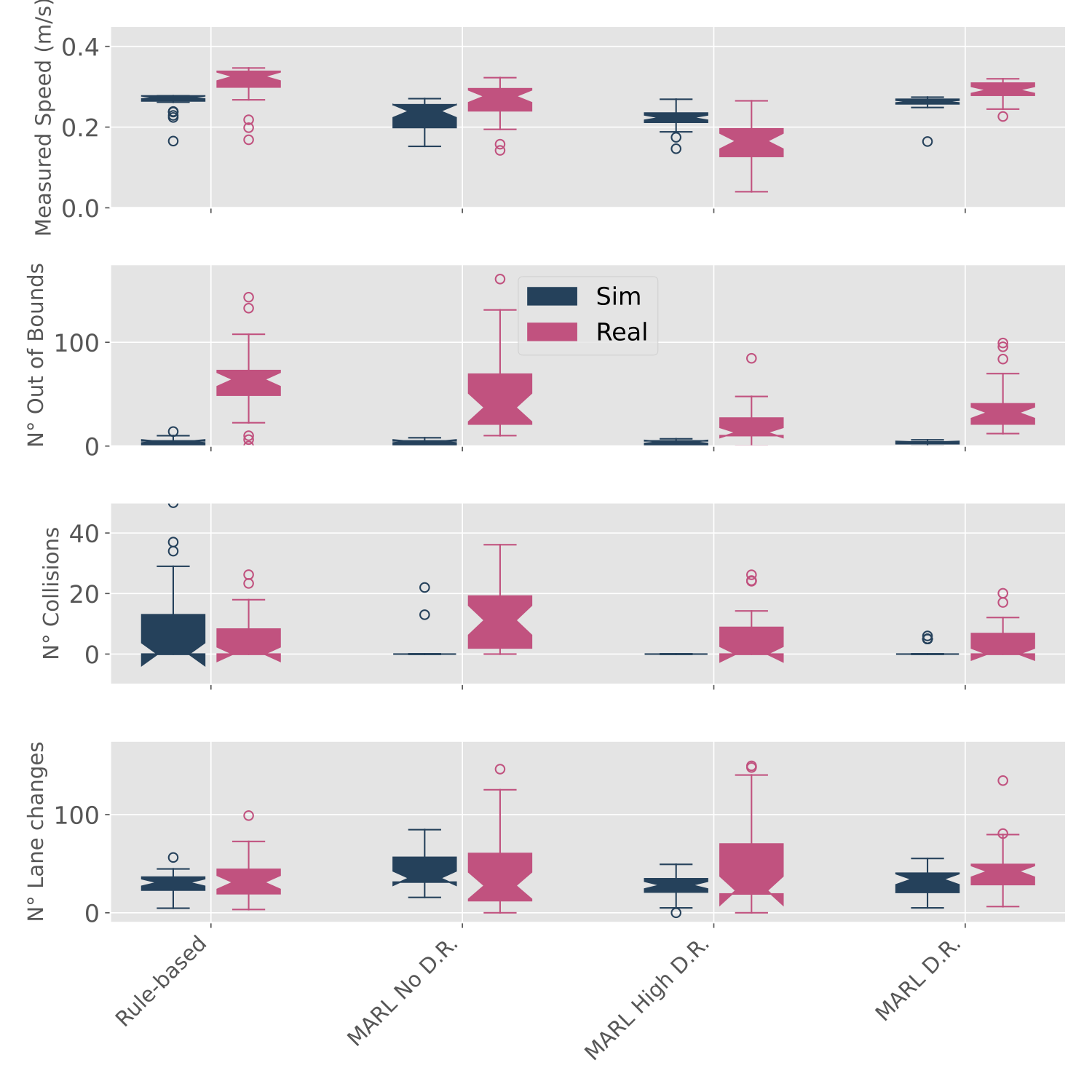}
    \caption{Distribution of performance metrics across different policies: 1) measured speed; 2) number of times a vehicle left the track; 3) number of collisions between vehicles; 4) number of times a vehicle changed lane. \\
    INTERPRETATION: The rule-based method tends to present the highest speed, track violations, and collisions; closely followed by the MARL policy with no D.R. In general, the policies trained with D.R. perform better in real life, this is partly because they change lanes significantly more times (compared to simulation) to avoid hazardous situations. The policy the medium D.R. (right-hand side policy) is the best one, followed by the one with high D.R. which tends to be more conservative in terms of speed -- leading to the least number of track exits and collisions, especially in real life.
    }
    \label{fig:results_detail}
\end{figure}

To better understand the resulting rewards of the policies, we analyze the four aspects that compose the reward function: speed, staying within the designated track, collisions, and lane changes. The numerical results of both the simulation and real tests are presented in \ref{fig:results_detail}. The average speed is the highest for the rule-based policy, closely followed by the med D.R. one. The policy with the slowest average measured speed is the one with high D.R. -- confirming that this is the most conservative policy, especially in real life. Similarly, for the times an agent goes outside the track bounds, the rule-based presents the highest number and the high D.R. the lowest; notice that this phenomenon happens mainly in real life and not in the simulated environment. For collisions, note that for all three MARL policies there are virtually no collisions in simulation, and that the policies trained with domain randomization tend to have lower collisions compared to the other policies. Finally, the average number of times an agent changes lanes is significantly larger in real life compared to simulation for policies trained with domain randomization, which can be interpreted as the agents choosing to switch lanes more to avoid the potential crashes that happen more often in real life.

\section{CONCLUSIONS} \label{sec:conclusions}

Autonomous Vehicles hold an enormous potential to improve safety and efficiency in various fields and applications. However, achieving reliable solutions will require solving the hard problem of coordination and collaboration of AVs. MARL is a tool that can help solve this problem, for it is an optimization-based method that can successfully allow agents to learn to collaborate by using shared observations and rewards. Nevertheless, the training of policies using MARL and other RL methods is usually heavily dependent on accurate simulation environments, which is hard to achieve due to \textit{reality gaps}. 

We present a method to train policies using MARL and to reduce the \textit{reality gap} when transferring them to the real world via adding domain randomization during training, which we show has a significant and positive impact in real performance compared to rule-based methods or policies trained without different levels of domain randomization. 

It is important to mention that despite the performance improvements observed when using domain randomization, its use presents diminishing returns as seen with the overly conservative policy, for it cannot completely close the \textit{reality gap} without increasing the fidelity of the simulator. Additionally, the amount of domain randomization to be used is case-specific and a theory for the selection of domain randomization remains an open question. The quantification and description of \textit{reality gaps} presents another opportunity for future research.






\section*{ACKNOWLEDGMENT}

This research was partially supported by the President's Scholarship Programme funded by Imperial College London, and the Chilean National Agency for Research and Development (ANID) through the "BECAS DOCTORADO EN EXTRANJERO" program, Grant No. 72210279.


\bibliographystyle{IEEEtran}
\bibliography{bib/marl,bib/traffic_management,bib/sim2real,bib/preliminaries,bib/results}

\end{document}